\documentclass[11pt]{article}
\usepackage{authblk}
\usepackage{coling2020}
\usepackage{times}
\usepackage{url}
\usepackage{xcolor}
\usepackage{latexsym}
\usepackage{graphicx}
\usepackage{multirow}
\usepackage{amssymb}
\usepackage{hyperref}

\usepackage{tabularx}
\usepackage{makecell}
\usepackage{xcolor}
\usepackage{tablefootnote}

\newcommand\legend[1]{\fcolorbox{white}{#1}{\rule{0pt}{4pt}\rule{4pt}{0pt}}}

\definecolor{synonym}{HTML}{b6d7a8}
\definecolor{cohyponym}{HTML}{38761d}
\definecolor{cohyponym3}{HTML}{6aa84f}
\definecolor{target}{HTML}{04df55}
\definecolor{transitivehypernym}{HTML}{1155cc}
\definecolor{transitivehyponym}{HTML}{c9daf8}
\definecolor{directhypernym}{HTML}{3c78d8}
\definecolor{directhyponym}{HTML}{a4c2f4}
\definecolor{unknownrelation}{HTML}{ea9999}
\definecolor{unknownword}{HTML}{e06666}
\definecolor{multiword}{HTML}{808080}
\definecolor{nopath}{HTML}{ed3833}

\newcolumntype{H}{>{\setbox0=\hbox\bgroup}c<{\egroup}@{}}

\newcommand{\specialcell}[2][l]{%
  \begin{tabular}[#1]{@{}c@{}}#2\end{tabular}}

\newcommand{\maskend}{\rule{0.3cm}{0.6pt}}
\newcommand{\maskmid}{\rule{0.3cm}{0.6pt} }

\usepackage{microtype}

\newenvironment{itemize2}
    {\begin{itemize}
        \vspace{-0.3em}
        \setlength{\abovedisplayskip}{0pt}
        \setlength{\belowdisplayskip}{0pt}
        \setlength{\itemsep}{5pt}
        \setlength{\parskip}{0pt}
        \setlength{\parsep}{0pt}
        \setlength{\topsep}{0pt}
        \setlength{\partopsep}{0pt}
    }
    {\vspace{-0.3em}
    \end{itemize}}

\newenvironment{enumerate2}
    {\begin{enumerate}
        \vspace{-0.05em}
        \setlength{\abovedisplayskip}{0pt}
        \setlength{\belowdisplayskip}{0pt}
        \setlength{\itemsep}{5pt}
        \setlength{\parskip}{0pt}
        \setlength{\parsep}{0pt}
        \setlength{\topsep}{0pt}
        \setlength{\partopsep}{0pt}
    }
    {\vspace{-0.08em}
    \end{enumerate}}

\colingfinalcopy 

\title{Always Keep your Target in Mind: Studying Semantics and \\ Improving Performance of Neural Lexical Substitution }

\author[$1,2,3$]{\textbf{Nikolay Arefyev}}
\author[$1,2$]{\textbf{Boris Sheludko}}
\author[$1$\thanks{~~Left Samsung}]{\textbf{Alexander Podolskiy}}
\author[$4$]{\textbf{Alexander Panchenko}}

\affil[$1$]{Samsung Research Center Russia, Moscow, Russia}
\affil[$2$]{Lomonosov Moscow State University, Moscow, Russia}
\affil[$3$]{HSE University, Moscow, Russia}
\affil[$4$] {Skolkovo Institute of Science and Technology, Moscow, Russia}
\affil[ ]{\texttt{narefjev@cs.msu.ru}}
\affil[ ]{\texttt{\{b.sheludko,a.podolskiy\}@samsung.com}}
\affil[ ]{\texttt{a.panchenko@skoltech.ru}}

\date{}

\begin{document}
\maketitle
\vspace{30pt}

\begin{abstract}
Lexical substitution, i.e. generation of plausible words that can replace a particular target word in a given context, is an extremely powerful technology that can be used as a backbone of various NLP applications, including word sense induction and disambiguation, lexical relation extraction, data augmentation, etc. In this paper, we present a large-scale comparative study of lexical substitution methods employing both rather old and most recent language and masked language models (LMs and MLMs), such as context2vec, ELMo, BERT, RoBERTa, XLNet. We show that already competitive results achieved by SOTA LMs/MLMs can be further substantially improved if information about the target word is injected properly. Several existing and new target word injection methods are compared for each LM/MLM using both intrinsic evaluation on lexical substitution datasets and extrinsic evaluation on word sense induction (WSI) datasets. On two WSI datasets we obtain new SOTA results. Besides, we analyze the types of semantic relations between target words and their substitutes generated by different models or given by annotators.
\end{abstract}

\section{Introduction}
%
%
\blfootnote{
    %
    %
    %
    
    \hspace{-0.65cm}  
    This work is licensed under a Creative Commons
    Attribution 4.0 International Licence.
    Licence details:
    \url{http://creativecommons.org/licenses/by/4.0/}.
    
    %
}

Lexical substitution is the task of generating words that can replace a given word in a given textual context. For instance, in the sentence ``\textit{My daughter purchased a new car}'' the word \textit{car} can be substituted by its synonym \textit{automobile}, but also with co-hyponym \textit{bike}, or even hypernym \textit{motor vehicle} while keeping the original sentence grammatical. Lexical substitution has been proven effective in various applications, such as word sense induction~\cite{amrami-2018}, lexical relation extraction~\cite{schick2019rare}, paraphrase generation, text simplification, textual data augmentation, etc. Note that the preferable type (e.g., synonym, hypernym, co-hyponym, etc.) of generated substitutes depends on the task at hand.

The new generation of language and masked language models (LMs/MLMs) based on deep neural networks enabled a profound breakthrough in almost all NLP tasks. These models are commonly used to perform pre-training of deep neural networks, which are then fine-tuned to some final task different from language modeling. 
However, in this paper we study how the progress in unsupervised pre-training over the last five years affected the quality of lexical substitution. We adapt context2vec~\cite{c2v}, ELMo~\cite{peters-etal-2018-deep}, BERT~\cite{devlin2018pretraining}, RoBERTa~\cite{Liu2019RoBERTaAR} and XLNet~\cite{yang2019xlnet} to solve lexical substitution task without any fine-tuning, but using additional techniques to ensure similarity of substitutes to the target word, which we call target word injection techniques.
We provide the first large-scale comparison of various neural LMs/MLMs with several target word injection methods on lexical substitution and WSI tasks. Our research questions are the following (i) which pre-trained models are the best for substitution in context, (ii) additionally to pre-training larger models on more data, what are the other ways to improve lexical substitution, and (iii) what are the generated substitutes semantically. More specifically, the main contributions of the paper are as follows\footnote{The repository for this paper: \url{https://github.com/bsheludko/lexical-substitution}}:

\begin{itemize2}
\item A comparative study of five neural LMs/MLMs applied for lexical substitution based on both intrinsic and extrinsic evaluation. 
\item A study of methods of target word injection for further lexical substitution quality improvement. 
\item An analysis of types of semantic relations (synonyms, hypernyms, co-hyponyms, etc.) produced by neural substitution models as well as human annotators. 
\end{itemize2}

\section{Related Work}
\label{sec:rel_work}

Solving the lexical substitution task requires finding words that are both appropriate in the given context and related to the target word in some sense (which may vary depending on the application of generated substitutes). To achieve this, unsupervised substitution models heavily rely on distributional similarity models of words (DSMs) and language models (LMs). Probably, the most commonly used DSM is \textit{word2vec} model~\cite{word2vec}. It learns word embeddings and context embeddings to be similar when they tend to occur together, resulting in similar embeddings for distributionally similar words. Contexts are either nearby words or syntactically related words \cite{levy-goldberg-2014-dependency}. In \cite{melamud-etal-2015-simple} several metrics for lexical substitution were proposed based on embedding similarity of substitutes both to the target word and to the words in the given context. Later \cite{pic} improved this approach by switching to dot-product instead of cosine similarity and applying an additional trainable transformation to context word embeddings. 

A more sophisticated \textit{context2vec} model producing embeddings for a word in a particular context (contextualized word embeddings) was proposed in \cite{c2v} and was shown to outperform previous models in a ranking scenario when candidate substitutes are given. The training objective is similar to {word2vec}, but context representation is produced by two LSTMs (a forward and a backward for the left and the right context), in which final outputs are combined by feed-forward layers. For lexical substitution, candidate word embeddings are ranked by their similarity to the given context representation. A similar architecture consisting of a forward and a backward LSTM is employed in {ELMo} \cite{peters-etal-2018-deep}. However, in {ELMo} each LSTM was trained with the LM objective instead. To rank candidate substitutes using {ELMo} \cite{soler-etal-2019-comparison} proposed calculating cosine similarity between contextualized {ELMo} embeddings of the target word and all candidate substitutes. This requires feeding the original example with the target word replaced by one of the candidate substitutes at a time. The average of outputs at the target timestep from all {ELMo} layers performed best. However, they found {context2vec} performing even better and explained this by the negative sampling training objective, which is more related to the task.   

Recently, Transformer-based models pre-trained on huge corpora with LM or similar objectives have shown SOTA results in various NLP tasks. BERT \cite{devlin2018pretraining} and RoBERTa~\cite{Liu2019RoBERTaAR} were trained to restore a word replaced with a special \texttt{[MASK]} token given its full left and right contexts (masked LM objective), while {XLNet} \cite{yang2019xlnet} predicted a word at a specified position given only some randomly selected words from its context (permutation LM objective). In \cite{zhou-etal-2019-bert}, {BERT} was reported to perform poorly for lexical substitution (which is contrary to our experiments), and two improvements were proposed to achieve SOTA results using it. Firstly, dropout is applied to the target word embedding before showing it to the model. Secondly, the similarity between the original contextualized representations of the context words and their representations after replacing the target by one of the possible substitutes are integrated into the ranking metric to ensure minimal changes in the sentence meaning. This approach is very computationally expensive, requiring calculation of several forward passes of BERT for each input example, depending on the number of possible substitutes. We are not aware of any work applying {XLNet} for lexical substitution, but our experiments show that it outperforms {BERT} by a large margin.

Supervised approaches to lexical substitution include \cite{szarvas-etal-2013-supervised,szarvas-etal-2013-learning,hintz-biemann-2016-language}. These approaches rely on manually curated lexical resources like WordNet, so they are not easily transferable to different languages, unlike those described above. Also, the latest unsupervised methods were shown to perform better~\cite{zhou-etal-2019-bert}. 

\section{Neural Language Models for Lexical Substitution with Target Word Injection}
\label{sec:methods}

To generate substitutes, we introduce several substitute probability estimators, which are models taking a text fragment and a target word position in it as input and producing a list of substitutes with their probabilities. To build our substitute probability estimators we employ the following LMs/MLMs: context2vec~\cite{c2v}, ELMo~\cite{peters-etal-2018-deep}, BERT~\cite{devlin2018pretraining}, RoBERTa~\cite{Liu2019RoBERTaAR} and XLNet~\cite{yang2019xlnet}. These models were selected to represent the progress in unsupervised pre-training with language modeling and similar tasks over the last five years.
Given a target word occurrence, the basic approach for models like context2vec and ELMo is to encode its context and predict the probability distribution over possible center words in this particular context. This way, the model does not see the target word. For MLMs, the same result can be achieved by masking the target word. This basic approach employs the core ability of LMs/MLMs of predicting words that fit a particular context. However, these words are often not related to the target. The information about the target word can improve generated substitutes, but what is the best method of injecting this information is an open question. 

\subsection{Target Word Injection Methods}
We experiment with several methods to introduce information about the original target word into neural lexical substitution models and show that their performance differs significantly. Suppose we have an example $LTR$, where $T$ is the target word, and $C = (L, R)$ is its context (left and right correspondingly). For instance, the occurrence of the target word \textit{fly} in the sentence ``\textit{Let me fly away!}'' will be represented as $T=$\textit{``fly''}, $L=$\textit{``Let me''}, $R=$\textit{``away!''}.

\paragraph{+embs} This method combines a distribution provided by a context-based substitute probability estimator $P(s|C)$ with a distribution based on the proximity of possible substitutes to the target $P(s|T)$. The proximity is computed as the inner product between the respective embeddings, and the softmax function is applied to get a probability distribution. However, if we simply multiply these distributions, the second will have almost no effect because the first is very peaky. To align the orders of distributions, we use temperature softmax with carefully selected temperature hyperparameter:
 $   P(s|T) \propto \exp(\frac{\langle emb_s, emb_T\rangle}{\mathcal{T}}).$
The final distribution is obtained by the formula
   $ P(s|C, T) \propto \frac{P(s|C)P(s|T)}{P(s)^\beta}.$
For $\beta=1$, this formula can be derived by applying the Bayes rule and assuming conditional independence of $C$ and $T$ given $s$. Other values of $\beta$ can be used to penalize frequent words, more or less. Our current methods are limited to generating only substitutes from the vocabulary of the underlying LM/MLM. Thus, we take word or subword embeddings of the same model we apply the injection to. Other word embeddings like word2vec may perform better, but we leave these experiments for future work. 

Word probabilities $P(s)$ are retrieved from wordfreq library\footnote{\url{https://pypi.org/project/wordfreq}} for all models except ELMo. Following~\cite{ranlp2019}, for ELMo, we calculate word probabilities from word ranks in the ELMo vocabulary (which is ordered by word frequencies) based on Zipf-Mandelbrot distribution and found it performing better presumably due to better correspondence to the corpus ELMo was pre-trained on.
 
\paragraph{Dynamic patterns} Following the approach proposed in~\cite{amrami-2018}, we replace the target word $T$ by ``$T$ and \maskend'' (e.g. \textit{``Let me fly and \maskmid away!''}). Then some LM/MLM is employed to predict possible words at timestep ``\maskend''. Thus, dynamic patterns provide information about the target word to the model via Hearst-like patterns.

\paragraph{Duplicate input} This method duplicates the original example while hiding the target word (e.g., \textit{``Let me fly away! Let me \maskmid away!''}). Then possible words at timestep ``\maskend'' are predicted. It is based on our observation that Transformer-based MLMs are very good at predicting words from the context when they fit the specified timestep (copying) while still giving a high probability to their distributionally similar alternatives. 

\paragraph{Original input} For MLMs such as BERT and RoBERTa, instead of masking the target word, we can leave it intact. Thus, the model predicts possible words at the target position while receiving the original target at its input. We noticed that unlike duplicate input, in this case, the MLM often puts the whole probability mass to the original target and gives very small probabilities to other words making their ranking less reliable. For XLNet, we can use such attention masks that the context words can see the target word in the content stream. Thus, the content stream becomes a full self-attention layer and sees all words in the original example. We do not apply the original input technique with context2vec and ELMo since, for these models, there is no reasonable representation that can be used to predict possible words at some timestep while depending on the input at that timestep, at least without fine-tuning. For other models, this option significantly outperforms target word masking and does not require many additional efforts. Hence, if not specified otherwise, we use it in our experiments with pure BERT, RoBERTa, XLNet estimators, and in {+embs} method when estimating $P(s|C)$ with BERT and XLNet.
%
%
\subsection{Neural Language Models for Lexical Substitution}
Different LMs/MLMs are employed as described below to obtain context-based substitute probability distribution $P(s|C)$. For each of them, we experiment with different target injection methods.
\paragraph{C2V} Context2vec encodes left, and right context separately using its forward and backward LSTM layers correspondingly and then combines their outputs with two feed-forward layers producing the final full context representation. Possible substitutes are ranked by the dot product similarity of their embeddings and the context representation. We use the original implementation\footnote{\url{https://github.com/orenmel/context2vec}} and the weights\footnote{\url{https://u.cs.biu.ac.il/~nlp/resources/downloads/context2vec/}} pre-trained on ukWac dataset.

\paragraph{ELMo} To encode left and right context with ELMo, we use its forward and backward stacked LSTMs, which were pre-trained with LM objective. However, there are no pre-trained layers to combine their outputs. Thus, we obtain two independent distributions over possible substitutes: one given the left context $P(s|L)$, another given the right context $P(s|R)$. To combine these distributions we use BComb-LMs method proposed in \cite{ranlp2019}, which can be derived similarly to +embs method  describe above: $P(s|L,R) = \frac{P(s|L)P(s|R)}{P^{\gamma}(s)}$. The original version of ELMo described in~\cite{peters-etal-2018-deep} is used, which is the largest version pre-trained on 1B Word Corpus.

\paragraph{BERT/RoBERTa} By default, we give our full example without any masking as input and calculate the distribution over the model's vocabulary at the position of the first subword of the target word. We employ BERT-large-cased and RoBERTa-large models as the best-performing ones. 
Unlike BERT and XLNet, we found that RoBERTa with {+embs} injection method performs better if cosine similarity instead of dot-product similarity is used when estimating $P(s|T)$ and the target word is masked when estimating $P(s|C)$. Thus, in the following experiments, we use these choices by default for {RoBERTa+embs} model.

\paragraph{XLNet} By default, we use the XLNet-large-cased model with the original input, obtaining substitute probability distribution similarly to BERT. 
We found that for short contexts, XLNet performance degrades. To mitigate this problem, we prepend the initial context with some text ending with the end-of-document special token.

\section{Baseline Lexical Substitution Models}

Lexical substitution models described above are compared to the best previously published results, as well as our re-implementations of the following baseline models proposed in~\cite{pic}.

\paragraph{OOC}
Out of Context model ranks possible substitutes by their cosine similarity to the target word and completely ignores given context. Following \cite{pic} we use dependency-based embeddings\footnote{\url{http://www.cs.biu.ac.il/~nlp/resources/downloads/lexsub_embeddings}} released by \cite{melamud-etal-2015-simple}.

\paragraph{nPIC}
Non-parameterized Probability In Context model returns the product of two distributions measuring the fitness of a substitute to the context and to the target:
${nPIC}(s|T,C) = P(s|T)\times P_n(s|C)$, where 
$P(s|T) \propto \exp(\langle embs_s, embs_T \rangle)$
and     
$P_n(s|C) \propto \exp(\sum\limits_{c \in C} \langle embs_s, embs'_c \rangle)$.
Here $embs$ and $embs'$ are dependency-based word and context embeddings, and $C$ are those words that are directly connected to the target in the dependency tree. 

\section{Intrinsic Evaluation}
\label{sec:intrinsic_eval}

We perform an intrinsic evaluation of the proposed models on two lexical substitution datasets.

\subsection{Experimental Setup}
Lexical substitution task is concerned with finding appropriate substitutes for a target word in a given context. 
This task was originally introduced in SemEval 2007 Task 10 \cite{mccarthy-navigli-2007-semeval} to evaluate how distributional models handle polysemous words. In the lexical substitution task, annotators are provided with a target word and its context. Their task is to propose possible substitutes. Since there are several annotators, we have some weight for each possible substitute in each example, which is equal to the number of annotators provided this substitute. 

We rank substitutes for a target word in a context by acquiring probability distribution over vocabulary on the target position. 
Lexical substitution task comes with two variations: candidate ranking and all-words ranking. In candidate ranking task, models are provided with a list of candidates. Following previous works, we acquire this list by merging all gold substitutes of the target lemma 
over the corpus. We measure performance on this task with Generalized Average Precision (GAP) that was introduced in \cite{gap}. GAP is similar to Mean Average Precision, and the difference is in the weights of substitutes: the higher the weight of the word, the higher it should be ranked. 
Following \cite{melamud-etal-2015-modeling}, we discard all multi-word expressions from the gold substitutes and omit all instances that are left without gold substitutes. 

In the all-vocab ranking task, models are not provided with candidate substitutes. Therefore, it is much harder task than the previous one. Models shall give higher probabilities to gold substitutes than to all other words in their vocabulary usually containing hundreds of thousands of words. 
Following \cite{pic}, we calculate the precision of the top 1 and 3 predictions (P@1, P@3) as an evaluation metric for the all-ranking task. Additionally, we look at the recall of top 10 predictions (R@10).

The following lexical substitution datasets are used:

\textbf{SemEval 2007 Task 10} \cite{mccarthy-navigli-2007-semeval}  consists of 2010 sentences for 201 polysemous words, 10 sentences for each. Annotators were asked to give up to 3 possible substitutes.
    
\textbf{CoInCo} or Concepts-In-Context dataset~\cite{kremer-etal-2014-substitutes} consists of about 2500 sentences that come from fiction and news. In these sentences, each content word is a separate example, resulting in about 15K examples. Annotators provided at least 6 substitutes for each example.
    

\subsection{Results and Discussion}

\paragraph{Comparison to previously published results}
\begin{table}[!t]
\centering
\footnotesize 
\begin{tabular}{|l|c|c|c|c|c|c|}
\hline
\multirow{2}{*}{ \bf Model} & \multicolumn{3}{c|}{ \bf SemEval 2007} & \multicolumn{3}{c|}{ \bf CoInCo} \\ \cline{2-7} 
 & \bf  GAP & \bf  P@1 & \bf  P@3 & \bf  GAP & \bf  P@1 & \bf  P@3 \\ \hline
Transfer Learning~\cite{hintz-biemann-2016-language} & 51.9 & - & - & - & - & -  \\ \hline
PIC~\cite{pic} & 52.4 & 19.7 & 14.8 & 48.3 & 18.2 & 13.8  \\ \hline
Supervised Learning~\cite{szarvas-etal-2013-learning} & 55.0 & - & - & - & - & -  \\ \hline
Substitute vector~\cite{melamud-etal-2015-modeling} & 55.1 & - & - & 50.2 & - & -  \\ \hline
context2vec~\cite{c2v} & 56.0 & - & - & 47.9 & - & -  \\ \hline
BERT for lexical substitution 
~\cite{zhou-etal-2019-bert}
\tablefootnote{We could not reproduce the results of \cite{zhou-etal-2019-bert} and their code was not available. } 
& \textbf{60.5} & \textbf{51.1} & - & \textbf{57.6} & \textbf{56.3} & -  \\ \hline
\hline
{XLNet+embs} &  \bf 59.6 & \bf  49.5 & \bf  34.9 &  \bf 55.6 & \bf  51.5 & \bf  39.9  \\ \hline
\specialcell{XLNet+embs (w/o target exclusion)} & 59.6 & 0.4 & 26.0 & 53.5 & 2.5 & 30.0  \\ 
\hline
\specialcell{XLNet+embs (w/o lemmatization)} & 59.2 & 38.3 & 27.5 & 53.2 & 34.5 & 27.1  \\ \hline
\specialcell{XLNet+embs ($\mathcal{T} = 1.0)$} & 52.6 & 34.8 & 24.6 & 49.4 & 40.5 & 30.4  \\ \hline
\end{tabular}
\caption{Intrinsic evaluation of our best model and its variations on lexical substitution datasets. 
}
\label{tab:prevbest_res}
\end{table}

Table~\ref{tab:prevbest_res} contains metrics for candidate and all-vocab ranking tasks. We compare our best model (XLNet+embs) with the best previously published results presented in \cite{pic}, context2vec (c2v) model \cite{c2v} and BERT for lexical substitution presented in \cite{zhou-etal-2019-bert}. The proposed model outperforms solid models such as PIC, c2v, and substitute vector by a large margin on both ranking tasks. Nevertheless, \cite{zhou-etal-2019-bert} reported better results than XLNet+embs in both lexical substitution tasks. In \cite{zhou-etal-2019-bert}, authors add a substitute validation metric that measures the fitness of a substitute to a context. It is computed as the weighted sum of cosine similarities of contextualized representations of words in two sentence versions: original and one where the target word is replaced with the substitute. This technique substantially improves predictions. However, substitute validation requires additional forward passes, hence, increasing computational overhead. Our methods need only one forward pass. Our approach is orthogonal to the substitute validation. Thus, a combination of two methods can improve results further. It is worth mentioning that BERT and XLNet work on a subword level. Hence, their vocabularies are much smaller in size (~30K subwords) than those of ELMo (800K words) or C2V (180K words) and contain only a fraction of possible substitutes. Thus, these models can be significantly improved by generating multi-token substitutes.

Additionally, table~\ref{tab:prevbest_res} includes ablation analysis of our best model. Using ordinary softmax (which is equivalent to setting $\mathcal{T} = 1.0$) results in all metrics decreasing by a large margin. 
Also, post-processing of substitutes has a significant impact on all-words ranking metrics. Since LMs/MLMs generate grammatically plausible word forms and often generate the target word itself among top substitutes, additional lemmatization and target exclusion is required to match gold substitutes. 
In~\cite{pic}, the authors used the NLTK English stemmer to exclude all forms of the target word. In~\cite{c2v} NLTK WordNet lemmatizer is used to lemmatize only candidates.
For a fair comparison, the same post-processing is used for all models in the following experiments.



\paragraph{Re-implementation of the baselines}
\begin{table*}[!thb]
\footnotesize 
\begin{center}
\begin{tabular}{|l|c|c|c|c|c|c|c|c|}
\hline
\multirow{2}{*}{\bf Model} & \multicolumn{4}{c|}{\bf SemEval 2007} & \multicolumn{4}{c|}{\bf CoInCo} \\ \cline{2-9} 
            & \bf GAP  & \bf  P@1  & \bf  P@3  & \bf  R@10 & \bf  GAP  & \bf  P@1  & \bf  P@3  & \bf  R@10 \\ \hline

OOC & 44.65 & 16.82 & 12.83 & 18.36 & 46.31 & 19.58 & 15.03 & 12.99 \\ \hline
nPIC & 52.46 & 23.22 & 17.61 & 27.4 & 48.57 & 25.75 & 19.12 & 17.33 \\ \hline
\hline

C2V &  \bf 55.81 & 7.79 & 5.92 & 11.03 & 48.32 & 8.01 & 6.63 & 7.54 \\ 
C2V+embs & 53.40 &  \bf 28.01 & \bf 21.72 & \bf  33.02 &  \bf 50.73 & \bf  29.64 & \bf  24.0 & \bf  21.37 \\ \hline

ELMo & 53.63 & 11.73 & 8.59 & 13.93 &  49.47 & 13.66 & 10.87 & 11.34  \\ 
ELMo+embs &  \bf  54.15 & \bf  31.95 & \bf 22.20 & \bf  31.82 & \bf  52.22 & \bf  35.92 & \bf  26.6 & \bf  23.80 \\ \hline

BERT & \bf 54.40 & 38.34 & 27.71 & 39.72 & 50.50 & 42.56 & 32.64 & 28.63  \\ 
BERT+embs &   53.88 &  \bf 41.64 & \bf  30.57 & \bf  43.48 & \bf  50.85 & \bf  46.05 & \bf  35.63 & \bf  31.37 \\ \hline

RoBERTa & 56.73 & 32.00 & 24.35 & 36.89 & 50.63 & 34.77 & 27.15 & 25.12  \\ 
RoBERTa+embs &  \bf 58.83 & \bf  44.13 & \bf  31.67 & \bf  44.70 & \bf  54.68 & \bf  46.54 & \bf  36.33 & \bf  32.03 \\ \hline

XLNet & 59.10 & 31.70 & 22.80 & 34.90 & 53.39 & 38.16 & 28.58 & 26.46  \\ 
XLNet+embs & \underline{\textbf{59.62}} & \underline{\textbf{49.48}} & \underline{\textbf{34.88}} & \underline{\textbf{47.18}} & \underline{\textbf{55.63}} & \underline{\textbf{51.48}} & \underline{\textbf{39.91}} & \underline{\textbf{34.91}}  \\ \hline

\end{tabular}
\caption{Comparison of our models and re-implemented baselines with the same post-processing.}
\label{tab:eval_res}
\end{center}
\end{table*}

In Table~\ref{tab:eval_res}, we compare our models based on different LMs/MLMs with and without {+embs} injection technique. Remember that BERT, RoBERTa, and XLNet see the target even if {+embs} is not applied, thus, providing already strong baseline results.
All compared models, including re-implemented OOC and nPIC, employ the same post-processing consisting of substitute lemmatization followed by target exclusion. First, we notice that our best substitution models substantially outperform word2vec based PIC and OOC methods. For example, the XLNet+embs gives 2x better P@1 and P@3 than the baselines. This indicates that proposed models are better at capturing the meaning of a word in a context as such, providing more accurate substitutes. On the candidate ranking task bare C2V model outperforms ELMo and BERT based models, but it shows the lowest Precision@1. We note that {+embs} technique substantially improves the performance of all models in all-vocab ranking task, and also increases GAP for the majority of models. 

\paragraph{Injection of information about target word}

\begin{figure}[!h]
\begin{center}
\includegraphics[width=0.98\textwidth]{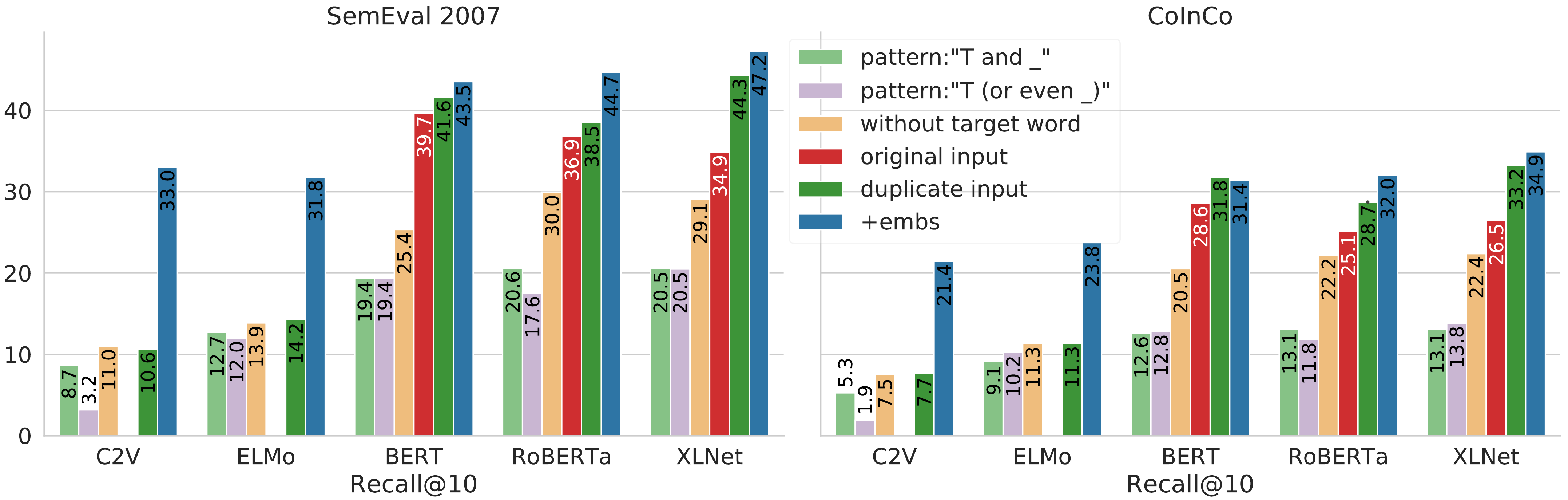}
\caption{Comparison of various target information injection methods (SemEval 2007 and CoInCo datasets).}
\label{fig:target-info}
\end{center}
\end{figure}

Next, we compare target injection methods described in Section~\ref{sec:methods}. Figure \ref{fig:target-info} presents the Recall@10 metric for all of our neural substitution models with each applicable target injection method. 

Application of dynamic patterns leads to lower performance even compared to the models that do not see the target word. 
Although we show the target word to the substitute generator, the pattern can spoil predictions, e.g., using ``T and \maskend'' pattern with a verb can produce words denoting subsequent actions related to the verb, but not its synonyms. When we use the original input without any masking, the models produce substitutes more related to the target, resulting in good baseline performance. 
Applying the {+embs} method leads to a significant increase of Recall@10 for all models, almost always being the best performing injection method. For C2V and ELMo, it gives 2-3 times better recall than all other injection methods. 
Duplicating input performs surprisingly good for Transformer-based models but does not help for LSTM-based C2V and ELMo. This is likely related to the previous observations that Transformers are very good at copying words from the surrounding context, thus, predicting the original target word, but also words with similar embeddings. The highest impact is for the XLNet model as it can not straightforwardly use information from the target position due to its autoregressive nature but can easily find the target in the sentence copy.
Overall, our experiments show that proper injection of the information about the target word results in much more plausible substitutes generated. 

\paragraph{Hyperparameters}
Since there is no development set of reasonable size in SemEval 2007 dataset, we decided to select hyperparameters for SemEval 2007 on CoInCo and vice versa. However, the selected hyperparameters turned out to be the same. Thus, for both datasets we use $\mathcal{T}{=}0.1$ for BERT+embs, XLNet+embs and ELMo+embs models, $\mathcal{T}{=}0.25$ for RoBERTa+embs and $\mathcal{T}{=}1.0$ for C2V+embs. For BERT+embs, XLNet+embs, RoBerta+embs, C2V+embs $\beta{=}0.0$. For ELMo and ELMo+embs $\gamma{=}0.5, \beta{=}1.5$.

\section{Extrinsic Evaluation}
\label{sec:extrinsic_eval}
In this section, we perform the extrinsic evaluation of our models applied to the Word Sense Induction (WSI) task. 
The data for this task commonly consists of a list of ambiguous target words and a corpus of sentences containing these words. Models are required to cluster all occurrences of each target word according to their meaning. Thus, the senses of all target words are discovered in an unsupervised fashion.  
For example, suppose that we have the following sentences with the target word \textit{bank}:
{ 
\begin{enumerate2}
    \item \textit{He settled down on the river \underline{bank} and contemplated the beauty of nature},
    \item \textit{They unloaded the tackle from the boat to the \underline{bank}}.
    \item \textit{Grand River \underline{bank} now offers a profitable mortgage}.
\end{enumerate2}
}

Sentences 1 and 2 shall be put in one cluster, while sentence 3 must be assigned to another. This task was proposed in several SemEval competitions \cite{agirre-soroa-2007-semeval,manandhar-etal-2010-semeval,jurgens-klapaftis-2013-semeval}. The current state-of-the-art approach \cite{amrami-2019} rely on substitute vectors, i.e., each word usage is represented as a substitute vector based on the most probable substitutes, then clustering is performed over these substitute vectors.
\begin{table}[!h]
\footnotesize
\centering
\begin{tabular}{|c|c|c|}
\hline

\textbf{Model} & \textbf{SemEval-2010 (AVG)} & \textbf{SemEval-2013 (AVG)} \\
\hline
\cite{amrami-2018}      & --    & 25.43$\pm$0.48 \\     
\hline
\cite{amrami-2019}      & \textbf{53.6}$\pm$1.2     & \textbf{37.0}$\pm$0.5      \\
\hline \hline

{C2V  }                      & \bf 38.9 \bf      & 18.2       \\
{C2V+embs }                  & 28.5      & \bf 21.7 \bf       \\

\hline
ELMo                        & 41.8      & 27.6       \\
ELMo+embs                        & \bf 45.3 &  \bf  28.2       \\

\hline
BERT                        & 52.0      & 34.5       \\
BERT+embs                        & \bf 53.8      & \bf 36.8       \\

\hline
RoBERTa & 49.6 & 34.0 \\
RoBERTa+embs & \bf 51.4 & \bf 34.5 \\

\hline
XLNet                        & 52.2      & 33.4       \\
XLNet+embs                        &  \bf  \underline{54.2}      &  \bf  \underline{37.3}    \\  
\hline
\end{tabular}
\caption{Extrinsic evaluation on word sense induction datasets. 
}
\label{tab:wsi-table}
\end{table}

We implemented a WSI algorithm using lexical substitutes from our models. The algorithm is a simplified version of methods described in \cite{amrami-2019,ranlp2019}. In the first step, we generate substitutes for each example, lemmatize them and take 200 most probable ones. We treat these 200 substitutes as a document. Then, TF-IDF vectors for these documents are calculated and clustered using agglomerative clustering with average linkage and cosine distance. The number of clusters maximizing the silhouette score is selected for each word individually.

In table~\ref{tab:wsi-table} we compare our lexical substitution models on two WSI datasets. For the previous SOTA models, which are stochastic algorithms, the mean and the standard deviation are reported. Our WSI algorithm is deterministic; hence, we report the results of a single run. Our best model achieves higher metrics than the previous SOTA on both datasets, however, the difference is within one standard deviation.
Similarly to the intrinsic evaluation results, our {+embs} injection method substantially improves the performance of all models for WSI, except for the C2V+embs model on SemEval-2010, which probably used suboptimal hyperparameters. 

\paragraph{Hyperparameters}
The optimal hyperparameters for WSI models were selected on the TWSI dataset~\cite{biemann-2012-turk}. For both evaluation datasets we used $\beta{=}2.0$ for BERT+embs, XLNet+embs, RoBERTa+embs and ELMo+embs models, $\beta{=}0.0$ for C2V+embs, $\mathcal{T}{=}2.5$ for BERT+embs, $\mathcal{T}{=}1.0$ for XLNet+embs, $\mathcal{T}{=}10.0$ for RoBERTa+embs, $\mathcal{T}{=}0.385$ for ELMo+embs and $\mathcal{T}{=}1.0$ for C2V+embs.

\section{Analysis of Semantic Relation Types}

In this section we analyze the types of substitutes produced by different neural substitution models.

\subsection{Experimental Setup}
For this analysis, the CoInCo lexical substitution dataset~\cite{kremer-etal-2014-substitutes} described above is used. We employ WordNet~\cite{miller1995wordnet} to find the relationship between a target word and each generated substitute. First, from all possible WordNet synsets two synsets containing the target word and its substitute with the shortest path between them are selected. Then relation between these synsets is identified as follows. If there is a direct relation between the synsets, i.e. synonymy, hyponymy, hypernymy, or co-hyponymy with a common direct hypernym, we return this relation. Otherwise, we search for an indirect relation, i.e. transitive hyponymy or hypernymy, or co-hyponymy with a common hypernym at a distance of maximum 3 hops from each synset. We also introduce several auxiliary relations: \textit{unknown-word} -- the target or the substitute is not found among WordNet synsets with the required PoS, \textit{unknown-relation} -- the target and the substitute are in the same WordNet tree, but no relation can be assigned among those described above, \textit{no-path} -- the target and the substitute are in different trees. 


\begin{figure}[!h]
\begin{center}
\includegraphics[width=0.98\textwidth]{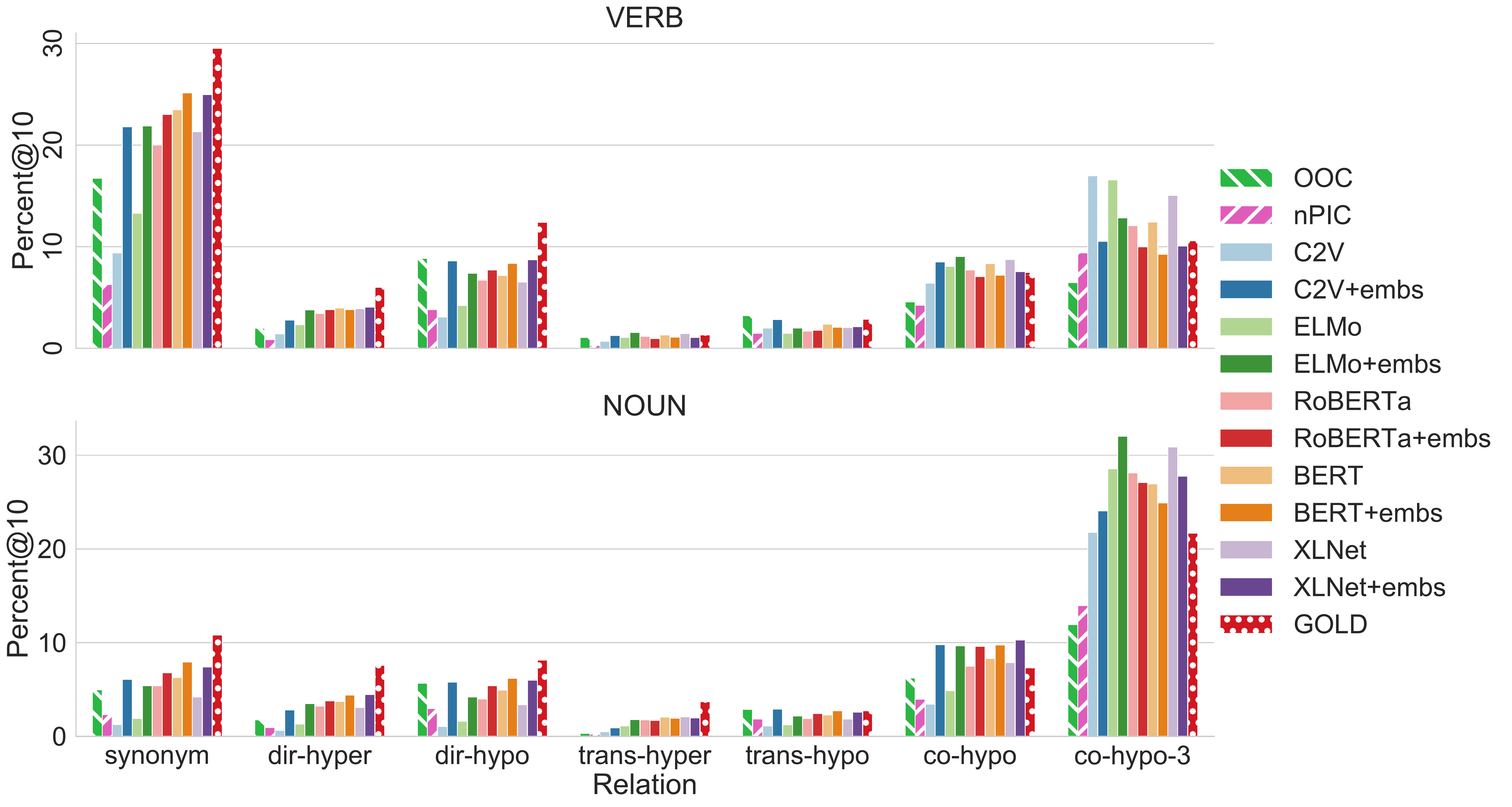}

\caption{Proportions of substitutes related to the target by various semantic relations according to WordNet. We took top 10 substitutes from each model and all substitutes from the gold standard.} 
\label{fig:relation-types}
\end{center}
\end{figure}

\subsection{Discussion of Results}

For nouns and verbs the proportions of non-auxiliary relations are shown in figure~\ref{fig:relation-types}, for all words and relations see Appendix~\ref{appendix:all_wordnet_relations}. 
Our analysis shows that a substantial fraction of substitutes has no direct relation to the target word in terms of WordNet, even in case of the gold standard substitutes. Besides, even human annotators occasionally provide substitutes of incorrect PoS, e.g., for {\it bright} as an adjective there is the verb {\it glitter} among gold substitutes. For adjectives and adverbs 18\% and 25\% of gold substitutes are unknown words (absent among synsets with the correct PoS), while for verbs and nouns less than 7\% are unknown. For baseline models OOC and nPIC, the overwhelming number of substitutes are unknown words. One of the reasons for this might be the fact that their vocabularies contain words with typos, but we also noticed that these models frequently do not preserve the PoS of the target word. 
The models based on LMs/MLMs produce much fewer unknown substitutes. Surprisingly, our +embs target injection method further reduces the number of such substitutes, achieving the proportion comparable to the gold standard. We can therefore suggest that our injection method helps to better preserve the correct PoS even for SOTA MLMs.

For both nouns and verbs, +embs target injection method consistently increases the proportions of synonyms, direct hyponyms and direct hypernyms, while decreasing the proportions of distantly related co-hyponyms (co-hypo-3) or unrelated substitutes. This is more similar to the proportions in human substitutes.
Thus, the addition of information from embeddings forces the models to produce words that are more closely related to the target word and more similar to human answers. For C2V and ELMo, which have no information on the target word, target word injection results in 2x-3x more synonyms generated.

\begin{figure*}[!h]
\centering
\includegraphics[width=0.98\textwidth]{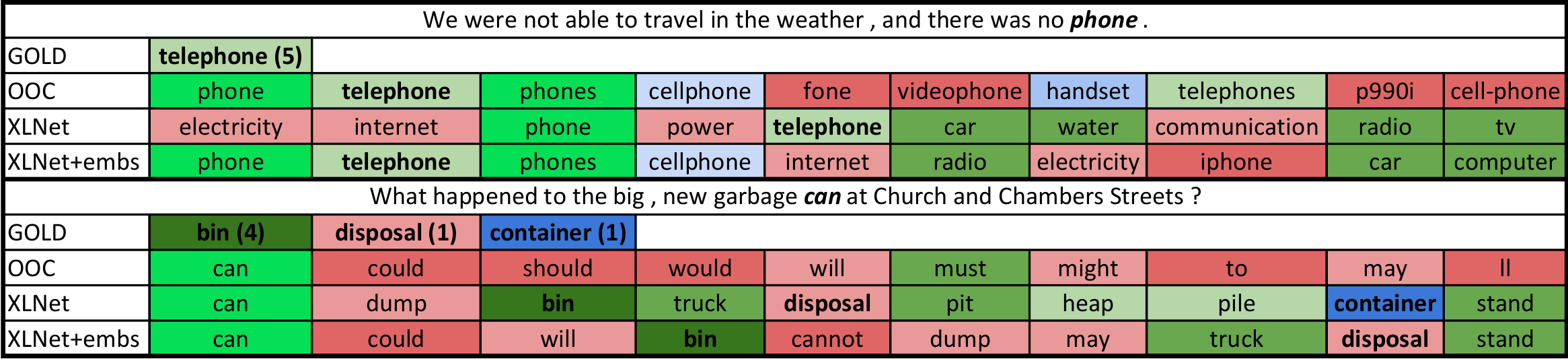}
\\{\scriptsize\textsf{Types of semantic relations: \legend{synonym}\,synonym \legend{cohyponym}\,co-hyponym \legend{cohyponym3}\,co-hyponym 3 \legend{target}\,target \legend{directhypernym}\,direct hypernym \legend{transitivehypernym}\,transitive hypernym \\\legend{directhyponym}\,direct hyponym \legend{transitivehyponym}\,transitive hyponym  \legend{unknownrelation}\,unknown-relation
\legend{unknownword}\,unknown-word
}}

\caption{Examples of top substitutes provided by annotators (GOLD), the baseline (OOC), and two presented models (XLNet and XLNet+embs). The target word in each sentence is in bold, true positives are in bold also. The weights of gold substitutes are given in brackets. Each substitute is colored according to its relation to the target word. Substitutes before post-processing are shown.}
\label{fig:interface}
\end{figure*}

For several sentences from SemEval 2007 dataset \cite{mccarthy-navigli-2007-semeval}, Figure~\ref{fig:interface} shows some examples of substitutes provided by the human annotators (GOLD) and generated by several models, see Appendix~\ref{appendix:subst_examples} for more examples and models. The first example shows the case when +embs injection method improves the result, ranking closely related substitutes, such as \textit{telephone}, \textit{cellphone}, higher. The substitutes provided by the bare XLNet model, such as \textit{electricity} and \textit{internet}, could be used in this context, but all the annotators had preferred the synonym \textit{telephone} instead. The second case is related to the failure of the proposed method. Bare XLNet model generated substitutes related to the correct sense of the ambiguous target word \textit{can}, and has all three gold substitutes among its top 10 predictions. In contrast, XLNet+embs produced words that are related to the most frequent sense: \textit{will}, \textit{could}, \textit{cannot}, etc. We hypothesize that this problem could potentially be alleviated by choosing individual temperature for each example based on the characteristics of the combined distributions; this is a possible direction for our further research.

\section{Conclusion}

We presented the first comparison of a wide range of LMs/MLMs with different target word injection methods on the tasks of lexical substitution and word sense induction. Our results are the following: (i) large pre-trained language models yield better results than previous unsupervised and supervised methods of lexical substitution; (ii) if properly done, the integration of information about the target word substantially improves the quality of lexical substitution models. The proposed target injection method based on a fusion of a context-based distribution $P(s|C)$ with a target similarity distribution $P(s|T)$ proved to be the best one. When applied to the XLNet model, it yields new SOTA results on two WSI datasets. Finally, we study the semantics of the produced substitutes. This information can be valuable for practitioners selecting the most appropriate lexical substitution method for a particular NLP application. 

\section*{Acknowledgements}

We thank the anonymous reviewers for the valuable feedback. The contribution of Nikolay Arefyev to the paper was partially done within the framework of the HSE University Basic Research Program funded by the Russian Academic Excellence Project ``5-100''.

\bibliographystyle{coling}
\bibliography{coling2020}

\begin{thebibliography}{}

\bibitem[\protect\citename{Agirre and Soroa}2007]{agirre-soroa-2007-semeval}
Eneko Agirre and Aitor Soroa.
\newblock 2007.
\newblock {S}em{E}val-2007 task 02: Evaluating word sense induction and
  discrimination systems.
\newblock In {\em Proceedings of the Fourth International Workshop on Semantic
  Evaluations ({S}em{E}val-2007)}, pages 7--12, Prague, Czech Republic, June.
  Association for Computational Linguistics.

\bibitem[\protect\citename{Amrami and Goldberg}2018]{amrami-2018}
Asaf Amrami and Yoav Goldberg.
\newblock 2018.
\newblock Word sense induction with neural bi{LM} and symmetric patterns.
\newblock In {\em Proceedings of the 2018 Conference on Empirical Methods in
  Natural Language Processing}, pages 4860--4867, Brussels, Belgium,
  October-November. Association for Computational Linguistics.

\bibitem[\protect\citename{Amrami and Goldberg}2019]{amrami-2019}
Asaf Amrami and Yoav Goldberg.
\newblock 2019.
\newblock Towards better substitution-based word sense induction.
\newblock {\em CoRR}, abs/1905.12598.

\bibitem[\protect\citename{Arefyev \bgroup et al.\egroup }2019]{ranlp2019}
Nikolay Arefyev, Boris Sheludko, and Alexander Panchenko.
\newblock 2019.
\newblock Combining lexical substitutes in neural word sense induction.
\newblock In {\em Proceedings of the International Conference on Recent
  Advances in Natural Language Processing (RANLP'19)}, RANLP~'19, pages 62--70,
  Varna, Bulgaria.

\bibitem[\protect\citename{Biemann}2012]{biemann-2012-turk}
Chris Biemann.
\newblock 2012.
\newblock Turk bootstrap word sense inventory 2.0: A large-scale resource for
  lexical substitution.
\newblock In {\em Proceedings of the Eighth International Conference on
  Language Resources and Evaluation ({LREC}'12)}, pages 4038--4042, Istanbul,
  Turkey, May. European Language Resources Association (ELRA).

\bibitem[\protect\citename{Devlin \bgroup et al.\egroup
  }2019]{devlin2018pretraining}
Jacob Devlin, Ming-Wei Chang, Kenton Lee, and Kristina Toutanova.
\newblock 2019.
\newblock {BERT}: Pre-training of deep bidirectional transformers for language
  understanding.
\newblock In {\em Proceedings of the 2019 Conference of the North {A}merican
  Chapter of the Association for Computational Linguistics: Human Language
  Technologies, Volume 1 (Long and Short Papers)}, pages 4171--4186,
  Minneapolis, Minnesota, June. Association for Computational Linguistics.

\bibitem[\protect\citename{Hintz and Biemann}2016]{hintz-biemann-2016-language}
Gerold Hintz and Chris Biemann.
\newblock 2016.
\newblock Language transfer learning for supervised lexical substitution.
\newblock In {\em Proceedings of the 54th Annual Meeting of the Association for
  Computational Linguistics (Volume 1: Long Papers)}, pages 118--129, Berlin,
  Germany, August. Association for Computational Linguistics.

\bibitem[\protect\citename{Jurgens and
  Klapaftis}2013]{jurgens-klapaftis-2013-semeval}
David Jurgens and Ioannis Klapaftis.
\newblock 2013.
\newblock {S}em{E}val-2013 task 13: Word sense induction for graded and
  non-graded senses.
\newblock In {\em Second Joint Conference on Lexical and Computational
  Semantics (*{SEM}), Volume 2: Proceedings of the Seventh International
  Workshop on Semantic Evaluation ({S}em{E}val 2013)}, pages 290--299, Atlanta,
  Georgia, USA, June. Association for Computational Linguistics.

\bibitem[\protect\citename{Kremer \bgroup et al.\egroup
  }2014]{kremer-etal-2014-substitutes}
Gerhard Kremer, Katrin Erk, Sebastian Pad{\'o}, and Stefan Thater.
\newblock 2014.
\newblock What substitutes tell us - analysis of an {``}all-words{''} lexical
  substitution corpus.
\newblock In {\em Proceedings of the 14th Conference of the {E}uropean Chapter
  of the Association for Computational Linguistics}, pages 540--549,
  Gothenburg, Sweden, April. Association for Computational Linguistics.

\bibitem[\protect\citename{Levy and
  Goldberg}2014]{levy-goldberg-2014-dependency}
Omer Levy and Yoav Goldberg.
\newblock 2014.
\newblock Dependency-based word embeddings.
\newblock In {\em Proceedings of the 52nd Annual Meeting of the Association for
  Computational Linguistics (Volume 2: Short Papers)}, pages 302--308,
  Baltimore, Maryland, June. Association for Computational Linguistics.

\bibitem[\protect\citename{Liu \bgroup et al.\egroup }2019]{Liu2019RoBERTaAR}
Yinhan Liu, Myle Ott, Naman Goyal, Jingfei Du, Mandar Joshi, Danqi Chen, Omer
  Levy, Mike Lewis, Luke Zettlemoyer, and Veselin Stoyanov.
\newblock 2019.
\newblock Roberta: A robustly optimized bert pretraining approach.
\newblock {\em ArXiv}, abs/1907.11692.

\bibitem[\protect\citename{Manandhar \bgroup et al.\egroup
  }2010]{manandhar-etal-2010-semeval}
Suresh Manandhar, Ioannis Klapaftis, Dmitriy Dligach, and Sameer Pradhan.
\newblock 2010.
\newblock {S}em{E}val-2010 task 14: Word sense induction {\&}disambiguation.
\newblock In {\em Proceedings of the 5th International Workshop on Semantic
  Evaluation}, pages 63--68, Uppsala, Sweden, July. Association for
  Computational Linguistics.

\bibitem[\protect\citename{McCarthy and
  Navigli}2007]{mccarthy-navigli-2007-semeval}
Diana McCarthy and Roberto Navigli.
\newblock 2007.
\newblock {S}em{E}val-2007 task 10: {E}nglish lexical substitution task.
\newblock In {\em Proceedings of the Fourth International Workshop on Semantic
  Evaluations ({S}em{E}val-2007)}, pages 48--53, Prague, Czech Republic, June.
  Association for Computational Linguistics.

\bibitem[\protect\citename{Melamud \bgroup et al.\egroup
  }2015a]{melamud-etal-2015-modeling}
Oren Melamud, Ido Dagan, and Jacob Goldberger.
\newblock 2015a.
\newblock Modeling word meaning in context with substitute vectors.
\newblock In {\em Proceedings of the 2015 Conference of the North {A}merican
  Chapter of the Association for Computational Linguistics: Human Language
  Technologies}, pages 472--482, Denver, Colorado, May{--}June. Association for
  Computational Linguistics.

\bibitem[\protect\citename{Melamud \bgroup et al.\egroup
  }2015b]{melamud-etal-2015-simple}
Oren Melamud, Omer Levy, and Ido Dagan.
\newblock 2015b.
\newblock A simple word embedding model for lexical substitution.
\newblock In {\em Proceedings of the 1st Workshop on Vector Space Modeling for
  Natural Language Processing}, pages 1--7, Denver, Colorado, June. Association
  for Computational Linguistics.

\bibitem[\protect\citename{Melamud \bgroup et al.\egroup }2016]{c2v}
Oren Melamud, Jacob Goldberger, and Ido Dagan.
\newblock 2016.
\newblock context2vec: Learning generic context embedding with bidirectional
  {LSTM}.
\newblock In {\em Proceedings of The 20th {SIGNLL} Conference on Computational
  Natural Language Learning}, pages 51--61, Berlin, Germany, August.
  Association for Computational Linguistics.

\bibitem[\protect\citename{Mikolov \bgroup et al.\egroup }2013]{word2vec}
Tomas Mikolov, Ilya Sutskever, Kai Chen, Greg Corrado, and Jeffrey Dean.
\newblock 2013.
\newblock Distributed representations of words and phrases and their
  compositionality.
\newblock {\em CoRR}, abs/1310.4546, august.

\bibitem[\protect\citename{Miller}1995]{miller1995wordnet}
George~A Miller.
\newblock 1995.
\newblock Wordnet: a lexical database for english.
\newblock {\em Communications of the ACM}, 38(11):39--41.

\bibitem[\protect\citename{Peters \bgroup et al.\egroup
  }2018]{peters-etal-2018-deep}
Matthew Peters, Mark Neumann, Mohit Iyyer, Matt Gardner, Christopher Clark,
  Kenton Lee, and Luke Zettlemoyer.
\newblock 2018.
\newblock Deep contextualized word representations.
\newblock In {\em Proceedings of the 2018 Conference of the North {A}merican
  Chapter of the Association for Computational Linguistics: Human Language
  Technologies, Volume 1 (Long Papers)}, pages 2227--2237, New Orleans,
  Louisiana, June. Association for Computational Linguistics.

\bibitem[\protect\citename{Roller and Erk}2016]{pic}
Stephen Roller and Katrin Erk.
\newblock 2016.
\newblock {PIC} a different word: A simple model for lexical substitution in
  context.
\newblock In {\em Proceedings of the 2016 Conference of the North {A}merican
  Chapter of the Association for Computational Linguistics: Human Language
  Technologies}, pages 1121--1126, San Diego, California, june. Association for
  Computational Linguistics.

\bibitem[\protect\citename{Schick and Sch{\"u}tze}2020]{schick2019rare}
Timo Schick and Hinrich Sch{\"u}tze.
\newblock 2020.
\newblock Rare words: A major problem for contextualized embeddings and how to
  fix it by attentive mimicking.
\newblock In {\em AAAI}, pages 8766--8774.

\bibitem[\protect\citename{Soler \bgroup et al.\egroup
  }2019]{soler-etal-2019-comparison}
Aina~Gar{\'\i} Soler, Anne Cocos, Marianna Apidianaki, and Chris
  Callison-Burch.
\newblock 2019.
\newblock A comparison of context-sensitive models for lexical substitution.
\newblock In {\em Proceedings of the 13th International Conference on
  Computational Semantics - Long Papers}, pages 271--282, Gothenburg, Sweden,
  23{--}27 May. Association for Computational Linguistics.

\bibitem[\protect\citename{Szarvas \bgroup et al.\egroup
  }2013a]{szarvas-etal-2013-supervised}
Gy{\"o}rgy Szarvas, Chris Biemann, and Iryna Gurevych.
\newblock 2013a.
\newblock Supervised all-words lexical substitution using delexicalized
  features.
\newblock In {\em Proceedings of the 2013 Conference of the North {A}merican
  Chapter of the Association for Computational Linguistics: Human Language
  Technologies}, pages 1131--1141, Atlanta, Georgia, June. Association for
  Computational Linguistics.

\bibitem[\protect\citename{Szarvas \bgroup et al.\egroup
  }2013b]{szarvas-etal-2013-learning}
Gy{\"o}rgy Szarvas, R{\'o}bert Busa-Fekete, and Eyke H{\"u}llermeier.
\newblock 2013b.
\newblock Learning to rank lexical substitutions.
\newblock In {\em Proceedings of the 2013 Conference on Empirical Methods in
  Natural Language Processing}, pages 1926--1932, Seattle, Washington, USA,
  October. Association for Computational Linguistics.

\bibitem[\protect\citename{Thater \bgroup et al.\egroup }2010]{gap}
Stefan Thater, Hagen F{\"u}rstenau, and Manfred Pinkal.
\newblock 2010.
\newblock Contextualizing semantic representations using syntactically enriched
  vector models.
\newblock In {\em Proceedings of the 48th Annual Meeting of the Association for
  Computational Linguistics}, pages 948--957, Uppsala, Sweden, July.
  Association for Computational Linguistics.

\bibitem[\protect\citename{Yang \bgroup et al.\egroup }2019]{yang2019xlnet}
Zhilin Yang, Zihang Dai, Yiming Yang, Jaime Carbonell, Russ~R Salakhutdinov,
  and Quoc~V Le.
\newblock 2019.
\newblock Xlnet: Generalized autoregressive pretraining for language
  understanding.
\newblock In {\em Advances in neural information processing systems}, pages
  5753--5763.

\bibitem[\protect\citename{Zhou \bgroup et al.\egroup
  }2019]{zhou-etal-2019-bert}
Wangchunshu Zhou, Tao Ge, Ke~Xu, Furu Wei, and Ming Zhou.
\newblock 2019.
\newblock {BERT}-based lexical substitution.
\newblock In {\em Proceedings of the 57th Annual Meeting of the Association for
  Computational Linguistics}, pages 3368--3373, Florence, Italy, July.
  Association for Computational Linguistics.

\end{thebibliography}

\newpage
\appendix

\section{Additional Examples of Lexical Substitutes}
\label{appendix:subst_examples}
\begin{figure*}[!h]
\centering
\includegraphics[width=0.98\textwidth]{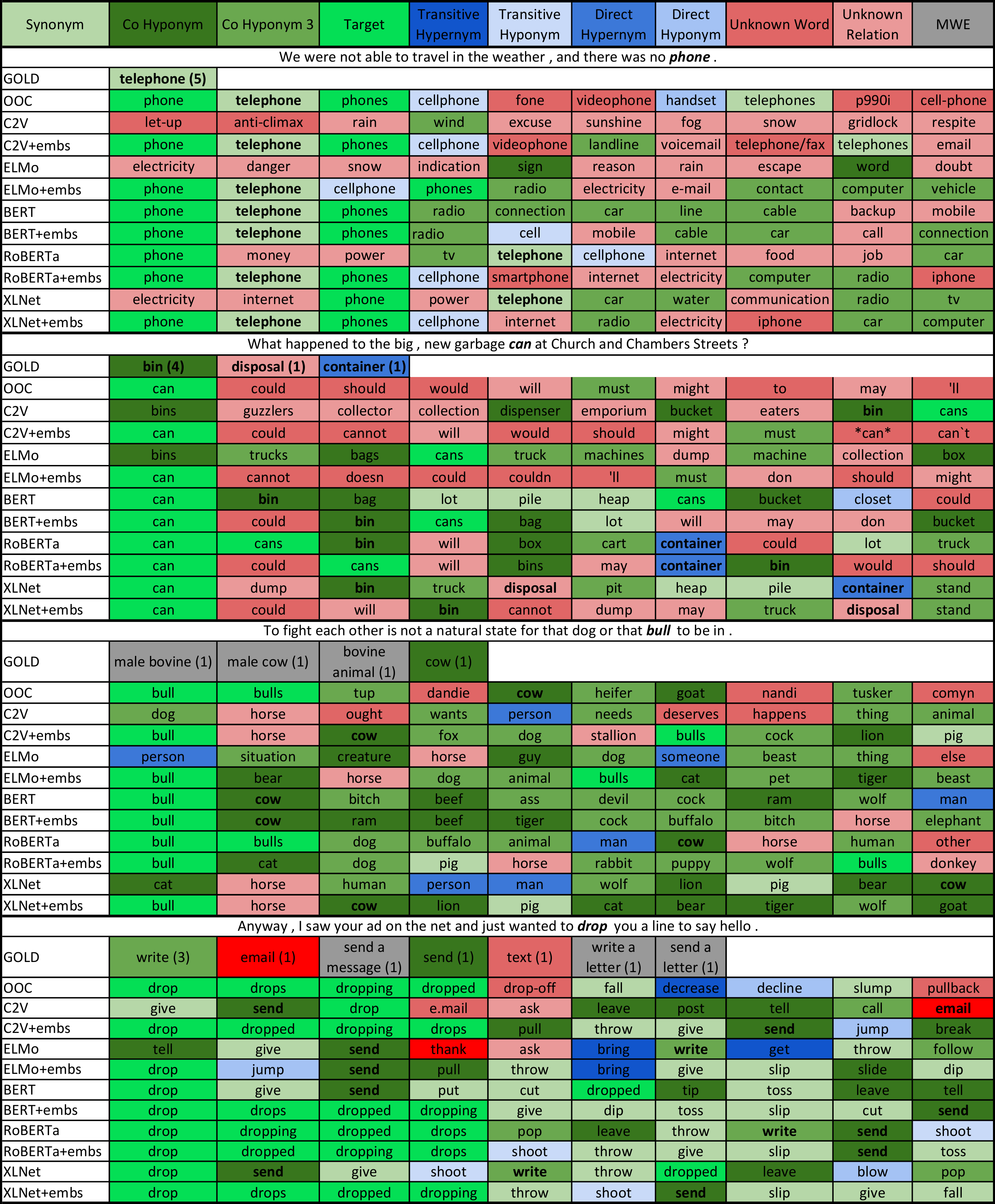}
\\{\scriptsize\textsf{Types of semantic relations: \legend{synonym}\,synonym \legend{cohyponym}\,co-hyponym \legend{cohyponym3}\,co-hyponym 3 \legend{target}\,target \legend{directhypernym}\,direct hypernym \legend{transitivehypernym}\,transitive hypernym \\\legend{directhyponym}\,direct hyponym \legend{transitivehyponym}\,transitive hyponym  \legend{unknownrelation}\,unknown-relation
\legend{unknownword}\,unknown-word
\legend{nopath}\,no-path
\legend{multiword}\,multiword expression
}}

\caption{Examples of substitutes produced by various lexical substitution methods based on original neural language models and their improved versions with +embs target word injection. Sentences are from SemEval 2007 dataset.}
\label{fig:additional}
\end{figure*}
\newpage
\section{Proportions of substitute types}
\label{appendix:all_wordnet_relations}

\begin{figure*}[!h]
\centering
\includegraphics[width=0.98\textwidth]{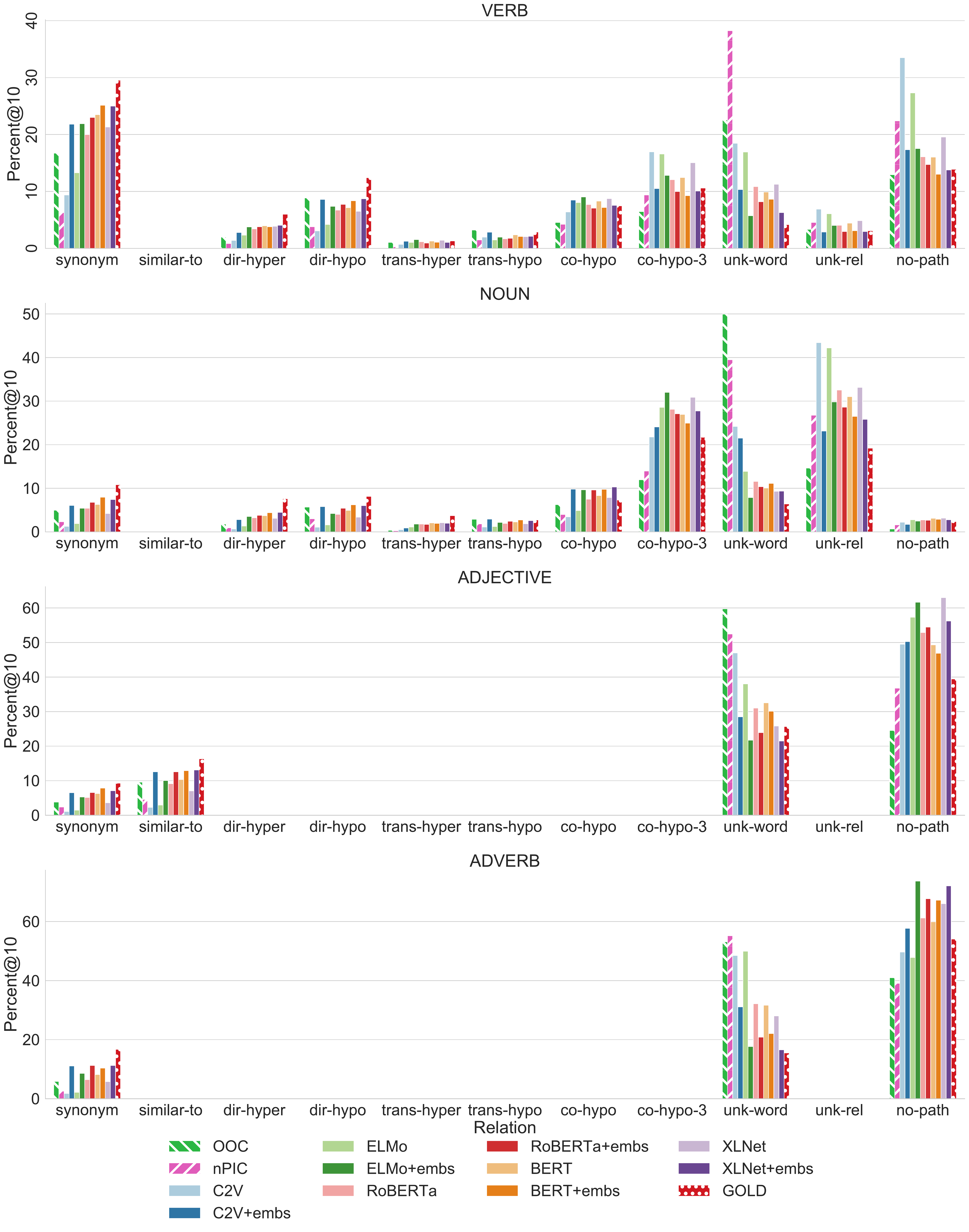}
\caption{Proportions of substitutes related to the target by various semantic relations according to WordNet. We took top 10 substitutes from each model and all substitutes from the gold standard. Examples are from the CoInCo dataset.}
\label{fig:additional_wordnet}
\end{figure*}

\end{document}